\documentclass{article}
\usepackage{spconf,amsmath,graphicx}
\usepackage{graphicx}
\usepackage{amssymb,amsmath,bm}
\usepackage{textcomp}
\usepackage{cite}
\usepackage{multirow}
\usepackage{float}
\usepackage{hyperref}
\usepackage{subfigure}
\title{Query-by-example Spoken Term Detection \\ using Attention-based Multi-hop Networks}
\name{Chia-Wei Ao and Hung-yi Lee\thanks{This work was sponsored
by  Ministry of Science and Technology, R.O.C.}}

\address{Graduate Institute of Communication Engineering, National Taiwan University\\
r04942094@ntu.edu.tw, hungyilee@ntu.edu.tw}

\begin{document}
\ninept
\maketitle
\begin{abstract}
Retrieving spoken content with spoken queries, or query-by-example spoken term detection (STD), is attractive because it makes possible the matching of signals directly on the acoustic level without transcribing them into text. Here, we propose an end-to-end query-by-example STD model based on an attention-based multi-hop network, whose input is a spoken query and an audio segment containing several utterances; the output states whether the audio segment includes the query. The model can be trained in either a supervised scenario using labeled data, or in an unsupervised fashion. In the supervised scenario, we find that the attention mechanism and multiple hops improve performance, and that the attention weights indicate the time span of the detected terms. In the unsupervised setting, the model mimics the behavior of DTW, and it performs as well as DTW but with a lower run-time complexity.
\end{abstract}
\section{Introduction}
\label{sec:intro}

Retrieving spoken content with spoken queries, also known as query-by-example spoken term detection (STD)~\cite{MetzeCSL14QbyE,MediaEval_QbyE_13,MediaEval_QbyE_14,szoke2015query,ButQUESST14,IS15}, is attractive because hand-held or wearable devices make spoken queries a natural choice.
%Because both the content and the queries are in the form of speech, it is possible to match the signals directly on the acoustic level, without first transcribing them into phonemes or words. 
%For low-resourced languages with scarce annotated data, or languages without written forms, recognition is difficult, so it makes sense to bypass the need for speech recognition, which usually involves learning from large quantities of annotated audio data.
The most intuitive way to search over spoken content for a spoken query is to directly match the audio signals to find those audio snippets that sound like the spoken query, and dynamic time warping (DTW)~\cite{zhang2009unsupervised} is widely used.  
%Since the audio events in speech signals can be produced at different speeds with different durations, the spoken content and the spoken query are generally not aligned at the same pace. 
%Dynamic time warping (DTW)~\cite{zhang2009unsupervised} is widely used in this case.  
Despite DTW's wide use, it has several drawbacks.
As typical DTW does not have trainable parameters, even in an online system that collects the training data from user feedback, the data cannot be directly used to improve the algorithm.
In addition, the time complexity of DTW is usually proportional to the product of the lengths of the spoken queries and audio segments, which for real applications is usually excessive.

Query-by-example STD by representing each word segment as a vector~\cite{levin2013fixed,SRAILICASSP15,kamper2015deep,chung2016audio,A2VmultiIS16} is much more efficient than the conventional Dynamic Time Warping (DTW) based approaches, because only the similarities between two single vectors are needed, in additional to the significantly better retrieval performance obtained~\cite{chung2016audio}. 
Audio segment representation is still an open problem.
Several approaches have been successfully used in STD~\cite{SRAILICASSP15,MyJournal_SVM,segment2vectorIS13,segment2vectorIS12}, but these approaches were developed primarily in more heuristic ways, rather than deep learning. 
  By learning RNN with an audio segment as the input and the corresponding word as the target, the outputs of the hidden layer at the last few time steps can be taken as the representation of the input segment~\cite{QbyELSTMICASSP15}.
  Audio segment embedding can also be jointly learned with their corresponding character sequences by multi-view approach~\cite{he2016multi}.  
Sequence-to-sequence Autoencoder is used to represent variable-length audio segments by vectors with fixed dimensionality, which is referred to as Audio Word2Vec~\cite{chung2016audio}. 
This previous approach assumes that speech segments to be retrieved have been pre-segmented at word boundaries, which is not realistic. 
However, it was shown that neural embeddings learned from pre-segmented audio can be applied for embedding arbitrary segments~\cite{settle2017query}. 
In this paper, we use attention mechanism to locate the time span of the input query in the utterances to be retrieved, so word boundary segmentation is not needed at both the training and testing stages, and attention is shown to improve the query-by-example performance.

The target of this paper is to develop an end-to-end deep learning model for query-by-example STD. 
On STD with text query, end-to-end approaches have been explored, in which  a function  which can map the acoustic features of an utterance  and a text query  to a confidence score is developed.
Along this direction, encouraging results have been obtained based on structured support vector machine (SVM)~\cite{STD_SVM_hidden_ICASSP13,SVM_keyword2,SVM_keyword1}. 
However, learning structured SVM  is computationally intensive, so this approach is hard to scale.
An end-to-end deep learning based system for text query  STD has been proposed~\cite{audhkhasi2017end}. 
Attention mechanism and multiple hops are not used in the model, which is different from the proposed approach in this paper.
 
%On the other hand, reasoning systems incorporating memory and attention mechanisms such as the memory network (MemNN)\cite{DBLP:journals/corr/WestonCB14,DBLP:journals/corr/SukhbaatarSWF15} and dynamic memory network (DMN)~\cite{DBLP:journals/corr/KumarISBEPOGS15} were shown to be very successful in end-to-end question answering (QA)~\cite{DBLP:journals/corr/BordesUCW15,DBLP:journals/corr/KumarISBEPOGS15,DBLP:journals/corr/XiongMS16,DBLP:journals/corr/HermannKGEKSB15,DBLP:journals/corr/ShihSH15}. 
%Some new attention mechanisms~\cite{DBLP:journals/corr/XiongZS16,DBLP:journals/corr/SeoKFH16} recently proposed are shown to be helpful on reading comprehension dataset where the answer to every question is a segment of text. 
%In vidual question answering (VQA), attention-based configurable convolutional neural network (ABC-CNN) can learn  question-guided attention~\cite{ABCCNN}, and it is shown that multiple hops yielded improved results compared to a single step~\cite{VQA_MultipleHop}.
%Attention-based multi-hop networks were also applied on machine comprehension of spoken content~\cite{Tseng}, and hierarchical attention model (HAM)~\cite{fang_hsu} further constructed tree-structured sentence representations for sentences from their parsing trees and estimate attention weights on different nodes of the hierarchies.

%proposed approach
In this paper, we propose an end-to-end query-by-example STD model.
The model is an attention-based multi-hop network, the input of which is a query and an audio segment (containing several utterances), and the output a confidence score representing whether the audio segment includes the query term.
In the network, the input spoken query is represented as a vector by an LSTM encoder.
We use the attention mechanism to locate the time span of the query term in the audio segment.
Similar to query expansion, multiple hops are used to update the spoken queries via information extracted from the audio segment.
Then a key term detector determines whether the query term exists in the input audio segment. 
These network components are all learned end-to-end, and the model can be learned in a supervised or unsupervised setting.
In a supervised setting, the model is learned from a set of labeled data, which can be collected by user feedback in real applications.
In an unsupervised setting, the neural network mimics the behavior of DTW, and it performs as well as DTW but with a lower run-time complexity.

\section{Framework and Training Scenario}
\label{sec:framework}

\begin{figure}[!th]
\centering
\includegraphics[scale=0.35]{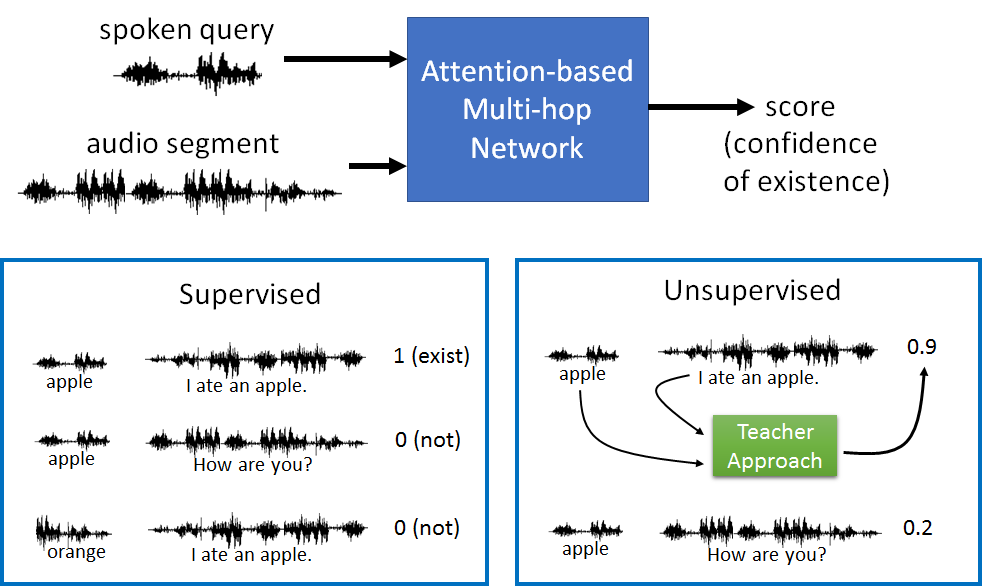} 
\caption{Framework and training scenarios}
\label{ch2_system}
\end{figure}

Shown in the upper half of Fig~\ref{ch2_system} is the framework of query-by-example STD using an end-to-end network.
%The network architecture is described in the next section.
The network input is the spoken query and an audio segment in the database to be retrieved.
Both the spoken query and the audio segment are represented by acoustic features such as MFCCs.
In this paper, the audio segments cover several utterances, and as such are much longer than the spoken queries.
The output of the network is a scalar.
The scalar represents the confidence that the  term in the spoken query exists in the audio segment.
Given a spoken query, the system ranks the audio segments in the database according to the confidence scores, and yields the search results.
It may seem that the proposed approach only extracts the target audio segments as opposed to locating the time spans of the query terms in the segments; in truth, the time spans are determined based on the attention mechanism in the network.
This is shown in the experimental results.

As shown in the lower half of Fig.~\ref{ch2_system}, the network is trained in two different scenarios. 
In the first scenario (left lower corner), labeled examples are collected, each pair of which is composed of a spoken query and an audio segment, including a label indicating whether the segment contains the term in the spoken query. 
In this case, network training is cast as a binary classification problem. %(containing the term or not).
Since labeled data is used in the first scenario, the network achieves better performance than query-by-example approaches such as DTW that use no labeled data.
In the second scenario (right lower corner), given a set of query-segment pairs, an existing query-by-example STD approach referred to here as the teacher approach (any method with good performance could be used here) assigns a score to each example pair.
This score represents the confidence that the segment includes the query.
The network thus learns to predict the confidence scores of the teacher approach given the same example pairs; this is hence a regression task.
In the second scenario, as no extra labeled data is needed, it is unsupervised.
Since the network is learned from an existing approach, it cannot outperform its teacher.
However, if the network performance is equivalent to that of the teacher approach, and if the network is faster than the teacher, it may be reasonable to use the network at testing time instead of the teacher approach. 

\section{Attention-based Multi-hop Network}\label{sec:approach}
In this section we describe the model architecture of the attention-based multi-hop network.

\begin{figure}[!th]
\centering
\includegraphics[scale=0.35]{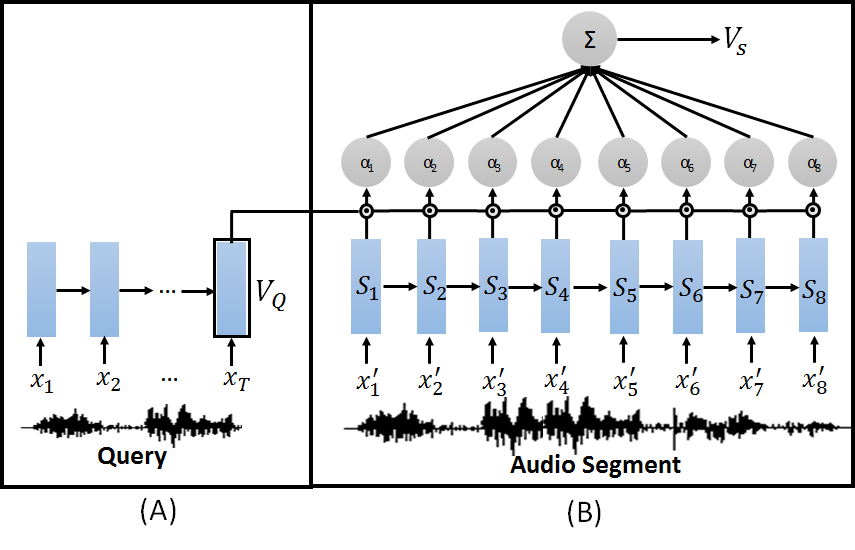} 
\caption{Attention mechanism}
\label{fig:att_model}
%\vspace{-3mm}
\end{figure}
%acoustic feature: 統一用 x
%document should be "audio segment" -- Lee

\subsection{Query Representation}\label{subsec:query}
Fig.~\ref{fig:att_model} (A) illustrates the encoding of the input spoken query into a vector representation $V_Q$.
The input query is a sequence of $T$ vectors, {$x_1,x_2,...,x_T$}, each vector $x_{i}$ of which is an acoustic feature vector such as MFCC.
In Fig.~\ref{fig:att_model} (A), a long short-term memory (LSTM) network~\cite{hochreiter1997long} takes one frame from the input spoken query sequentially at a time. 
After going through all the frames in the query, the query vector representation $V_{Q}$ is the hidden layer output of the LSTM network at the last time index.

%The hidden layer output of the forward LSTM (blue rectangle) at time index $t$ is denoted by $y^{f}_t$; that of the backward LSTM (pink rectangle) is $y^{b}_{t}$. 
%After going through all the frames in the query, the query vector representation $V_{Q}$, or $V_{Q} = [y^{f}_T \| y^{b}_{1}]$\footnote{Here the symbol [$\cdot$$\|$$\cdot$] denotes concatenation of two vectors.} is formed as the concatenation of the hidden layer output of the forward LSTM network at the last time index $y^{f}_T$ and that of the backward GRU network at the first time index $y^{b}_1$.

\subsection{Audio Segment Representation with Attention}\label{subsec:document}
Fig.~\ref{fig:att_model} (B) shows an audio segment (containing several utterances) in the database to be retrieved; although this is a lengthy acoustic feature sequence, we show only eight features for simplicity. %with many utterance?
The LSTM in Fig~\ref{fig:att_model} (B) goes through the whole document and encodes each frame\footnote{The LSTMs used in Figs~\ref{fig:att_model} (A) and (B) are the same.}.
The vector representation of the $t$-th frame $S_{t}$ is the hidden layer outputs of the LSTM network. %that is, $S_t = [y^{f}_{t} \| y^{b}_t]$.
This process can be completed off-line, before the spoken query is submitted.

Then the attention value $\alpha_t$ for each time index ${t}$ is the cosine similarity between the query vector $V_{Q}$ (obtained in Fig.~\ref{fig:att_model} (A)) and the vector representation $S_{t}$ of each frame, $\alpha_t = S_t \odot V_{Q}$, where symbol $\odot$ denotes cosine similarity between two vectors. 
We normalize attention values $\alpha_t$ as $\alpha_t^\prime$. %such that they sum to one over the whole story.
The score list is normalized using the softmax activation function:
     \begin{equation} 
		\alpha_{i}^\prime = \frac{exp(\alpha_{i})}{\sum_{i=1}^{T} exp(\alpha_{i})}  \label{sharp_norm}
\end{equation}
This has been widely used in many existing neural attention frameworks \cite{xu2015ask, xu2015show, bahdanau2014neural, weston2014memory}. %and works well with noisy data.
Then vectors $S_{t}$ from the LSTM network for every frame in the audio segment are weighted with this normalized attention value $\alpha_{t}^\prime$ and summed to yield the segment representation vector $V_{S} = \sum_{t}\alpha_{t}^{\prime}S_{t}$, which is used to determine the confidence score for spoken query $V_Q$.
To ensure a time complexity linear to the length of the input audio segment, we do not use more sophisticated attention models~\cite{seo2016bidirectional}; thus the approach is faster than DTW.
  
%\textbf{Smoothing}: The sharpening normalization method puts more focus on a single feature vector $S_{i}$, and can hence negatively affect the model's performance. We thus propose a way for the model to aggregate selections from multiple top-scored frames, thus bringing more diversity to the model by taking into account more input locations. We replace the exponential function in equation (\ref{sharp_norm}) with logistic sigmoid function $\sigma$:
%       \begin{equation} 
%		\alpha_{i}^\prime = \frac{\sigma(\alpha_{i})}{\sum_{i=1}^{T} \sigma(\alpha_{i})}  \label{smooth_norm}
%\end{equation}

\subsection{Hopping} \label{sec:hop}

\begin{figure}[!th]
\centering
\includegraphics[scale=0.3]{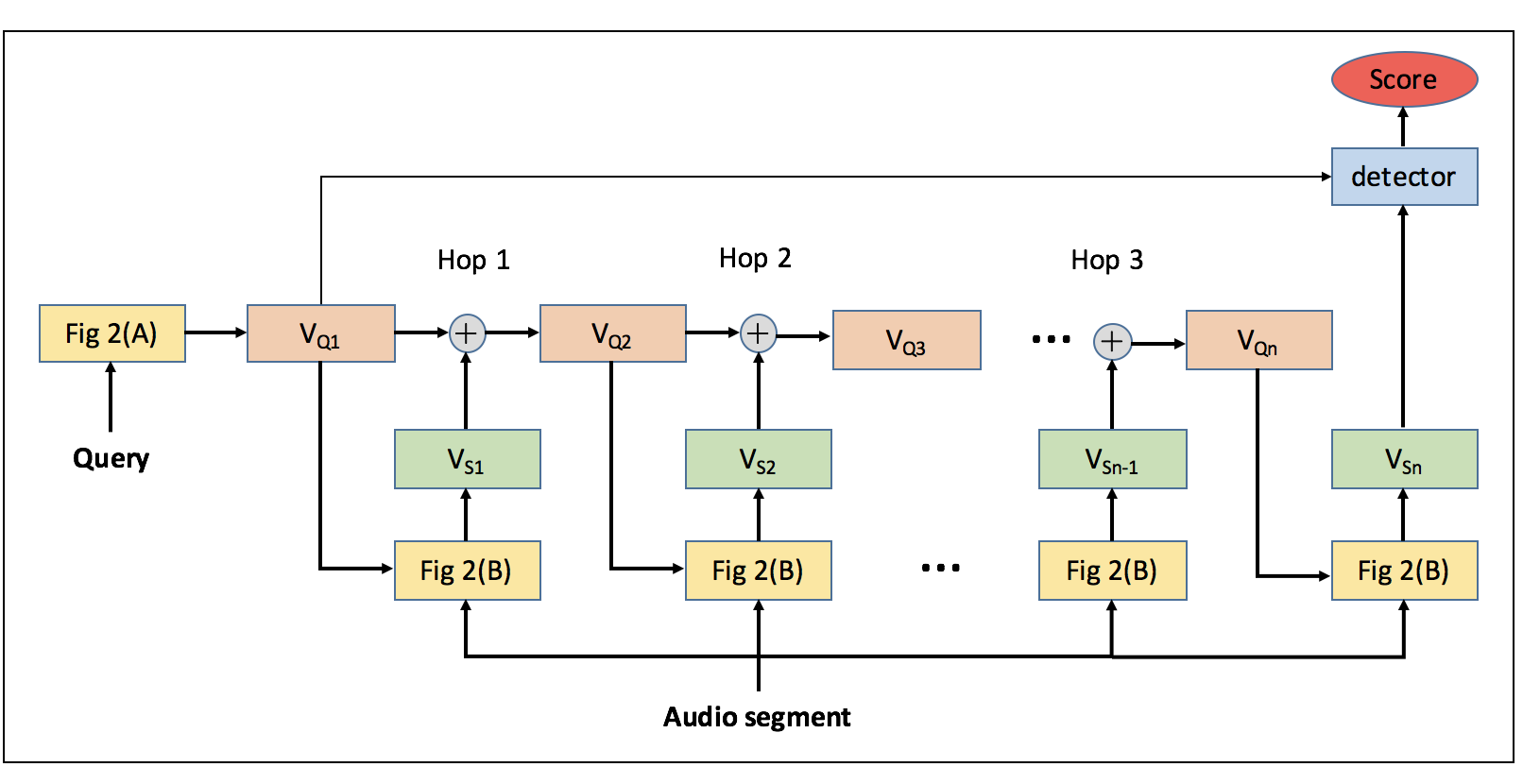} 
\caption{The hopping operation}
\label{fig:hopping}
\vspace{-2mm}
\end{figure}
%應該把 detector 加進去 
%把 classifier 改成 detector (因為在 unsupervised 的 case，並不是 detector) -- Lee
%Fig 3 (A), (B) -> Fig 2 -- Lee
%Document -> Audio segment -- Lee

Fig.~\ref{fig:hopping} illustrates hopping: the input spoken query is first converted into a vector $V_{Q_1}$ by the module in Fig.~\ref{fig:att_model} (A), after which the module in (B) uses this $V_{Q_1}$ to compute the attention values {$\alpha_{t}$} to obtain the story vector $V_{S_1}$.
Then $V_{Q_1}$ and $V_{S_1}$ are summed to form the new question vector $V_{Q_2}$.
This process is the first hop (hop 1).
The output of the first hop $V_{Q_2}$ can be used to compute the new attention values to obtain a new story vector $V_{S_2}$. 
This can be seen as \textit{relevance feedback}~\cite{MyJournal_SVM,RFICASSP13}, where the machine goes over the audio segment again, extracting information to expand the query to form a new query vector. 
Again, $V_{Q_1}$ and $V_{S_1}$ are summed to form $V_{Q_2}$ (hop 2).
After $n$ hops ($n$ is predefined), the output of the last hop $V_{S_n}$ is used to calculate the confidence score.

%Relevance feedback on STD:
%Enhanced Spoken Term Detection Using Support Vector Machines and Weighted Pseudo Examples, TASLP 2013.
%Using Parallel Tokenizers with DTW Matrix Combination for Low-Resource Spoken Term Detection, ICASSP 2013.

\subsection{Keyword Detection} \label{subsec:key_det}
Finally, as shown in the upper half of Fig \ref{fig:hopping}, a detector determines the confidence score based on query vector representation $V_Q$ and utterance vector representation $V_{S_n}$. 
Here we use three ways to calculate this score:
(1) Use the cosine similarity between $V_Q$ and $V_{S_n}$ as the score;
(2) Use the detector -- a connected feedforward neuron network -- taking $V_Q$ and $V_{S_n}$ as input, and output a scalar as the confidence score; 
(3) Combine approaches (1) and (2): a neural network takes as input the query vector $V_Q$, utterance vector $V_{S_n}$, and cosine similarity, and outputs a score.
% In the real application, the training examples and its supervised labels can be collected from user feedback

\section{Experimental Setup}
We used the LibriSpeech corpus~\cite{panayotov2015librispeech} as the data for the experiments.
To train the attention-based multi-hop network, some query-segment pairs were needed as training examples.
70,000 training examples were used in the experiments, including 500 different spoken queries; all audio segments were from the LibriSpeech \emph{train-clean-360} set.
In the supervised scenario, the label for each example (a query-segment pair) specified whether the audio segment contains the spoken query\footnote{This is easily determined using the manual transcriptions of the audio segments available from the LibriSpeech corpus.}.
In the unsupervised scenario, each example is labeled using the score from the DTW algorithm~\cite{dtwcode}. 
There are three testing sets. 
In all testing sets, the audio segments were from the \textit{train-other-500} set in LibriSpeech; thus the audio segments in the training and testing sets did not overlap.
As described below, the spoken queries were different.  %用哪一個set testing 呢? - Lee
\begin{itemize}
\item{Testing Set 1: There were 1,500 query-segment pairs, including 30 different spoken queries (each query has 50 examples in average). The spoken queries were all from the training set.} % There are xxxxx audio segments.??? 要寫嗎?
\item{Testing Set 2: As with testing set 1, this set also had 1,500 query-segment pairs with 30 different spoken queries (each query has 50 examples in average). The spoken queries in this set had the same text form as testing set 1, but did not come from the training set.} 
\item{Testing Set 3: This set had 10,000 query-segment pairs with 100 different spoken queries (each query has 100 examples in average). In this set, the spoken queries were not from the training set, and the text form of the spoken queries never appeared in the training queries.}	
\end{itemize}
All the spoken queries corresponds to single words, but the proposed approach can also be applied on phrases.
39-dimension MFCCs were used as the acoustic features.
Both attention-based multi-hop network and DTW baseline used the same set of features, so they can be fairly compared.
Mean average precision (MAP) was used as the evaluation measure. %for query-by-example STD.
%section 5, test set 1 and 2: do you mean "30 queries and 50 example pairs per query"?
%section 5, test set 3: do you mean "100 queries and 100 example pairs per query"?

    %Although the oracle word boundaries were used here for the query-by-example STD in the preliminary tests, the comparison in Section~\ref{sec:qbe_std} was fair since the baselines used the same segmentation.
    %SA and DSA were implemented with Theano~\cite{bergstra+al:2010-scipy,DBLP:journals/corr/abs-1211-5590}.
    The network structure and hyper-parameters were set as below without further tuning if not specified.  
     The LSTM encoder consisted of two hidden layers with 128 LSTM units.
        The networks were trained for 100 epochs using ADAM~\cite{kingma2014adam}. %without momentum, with a fixed learning rate of 0.01.
       The keyword detector was a network with four hidden layers with 128,  64, 32 and 2 neurons respectively.
 In the supervised scenario, the attention-based multi-hop network was a binary classifier trained using cross-entropy loss; in the unsupervised scenario, mean square error was the loss function.
     
% \begin{itemize}
%      \item{
%        The RNN encoder consisted of two hidden layers with 128 LSTM units. %That is, each input segment was mapped into a 128-dim vector representation.
%           }
%      \item{
%        The networks were trained for 100 epochs using ADAM~\cite{kingma2014adam} without momentum, with a fixed learning rate of 0.01.
%      }
%            \item{
%            Unless otherwise specified, the keyword detector was a network with four hidden layers with 128,  64, 32 and 2 neurons respectively. }
%%The keyword detectors were end-to-end learned with other part of the network.
%            \item{
%      In the supervised scenario, the attention-based multi-hop network was a binary classifier trained using cross-entropy loss; in the unsupervised scenario, mean square error was the loss function.
%      } %我這樣寫對嗎? -- Lee /
%    \end{itemize}
    
\section{Experimental Results}
In Sections~\ref{subsec:att_exp}, \ref{subsec:attana_exp} and \ref{subsec:hop_exp}, we consider the supervised scenario.
The results of the unsupervised scenario are presented in Section~\ref{subsec:un_exp}.

\subsection{Attention-based Model} \label{subsec:att_exp}

\begin{table}[h]
\vspace{-5mm}
	 \centering   
     \caption{Results of attention-based network with a single hop. Rows (A) and (B) are baselines.
     Part (C) is attention-based networks. 
     NN, Cos, and NN+Cos are respectively the three detectors from Section~\ref{subsec:key_det}.     }
	 \label{table:att_exp}
 %    \resizebox{\columnwidth}{!}{%
	 \begin{tabular}{|c|l|c|c|c|c|}
		 \hline
         \multicolumn{2}{|c|}{Approach}  & Test 1 &Test 2 & Test 3 \\
		 \hline \hline
		 \multicolumn{2}{|c|}{(A): DTW} & 0.6173 & 0.5778 & 0.5678\\
		 \hline     
         	 \multicolumn{2}{|c|}{(B): Network  without Attention} & 0.5935 & 0.5563 & 0.5468\\
         \hline 
         \hline
		(C): & (1) NN & 0.6523 & 0.6246 & 0.5754 \\
		 \cline{2-5} 
   Attention-based       & (2) Cosine  & 0.6331 & 0.6043 & 0.5746 \\
		 \cline{2-5}
Network & (3) NN+Cos & 0.6268 & 0.6370 & 0.5759 \\ 
		 \hline
      %   \hline
%         \multirow{2}{*}{(D): Network with attention} & (1)  NN &
%		 0.6237 & 0.5947 & 0.5628\\
%		 \cline{2-5}
%		 & (2)  Cosine & 0.6216 & 0.5844 & 0.5580 \\
%		 \cline{2-5}
%		  (smoothing)& (3)  NN+Cos&0.6203&0.5926	&0.5682 \\
%		 \hline
%        \hline
%		 \multirow{2}{*}{(E): BiLSTM} & (1)  NN &
%		 0.6355 & 0.6187 & 0.5664\\
%		 \cline{2-5}
%		 & (2)  Cosine& 0.6254 &0.6144 &0.5663 \\
%		 \cline{2-5}
%		 (sharpening)& (3)  NN+Cos &0.6192&0.6244 &0.5793 \\
%		 \hline
         \hline
		 \multirow{3}{*}{(D): (A)+(C)} & (1)  NN&
		 {0.6720} & 0.6340 & 0.5868\\
		 \cline{2-5}
		 & (2)  Cosine & 0.6433&0.6002&0.5843 \\
		 \cline{2-5}
		 & (3)  NN+Cos &0.6451&0.6309	&0.5808 \\
		 \hline
	   \end{tabular}%
  %     }    
\end{table}

Table~\ref{table:att_exp} shows the results of the attention-based model with a single hop. 
Rows (A) and (B) are two baselines:
Row (A) is the MAP of the search results ranked according to DTW similarities on the three testing sets, 
and row (B) is the model without attention mechanism. %The overall architecture is similar to the model we mentioned above. 
We used an  LSTM to encode both the spoken query and audio segment as a vector representation by taking the hidden layer output of the LSTM network at the last time index.
Next, we input to the neural network key term detector the query vector and audio segment representations. The detector then outputs a score representing the confidence that the query appears in the audio segment.
We find that without the attention mechanism, even though the networks are learned from labeled training data, they are outperformed by DTW, which needs no training data (rows (B) v.s. (A)).

We observe that the results of attention-based networks outperform those without attention and DTW (part (C) v.s. rows (A), (B)).
%This shows that the attention mechanism is helpful. 
From Table~\ref{table:att_exp}, we note that compared to DTW, the attention model yields larger improvements on test sets 1 and 2.
This shows that even though the training and testing queries are from different speakers, the attention-based model still learns the keyword acoustic patterns, which are speaker-independent.
%Despite speaker difference, the model still could do the retrieval well. The attention model has learned the keyword acoustic pattern which is independent of speaker. 
However, the attention-based network yields little improvement on test set 3; it is more difficult for the network to transfer what it has learned to keywords it has never seen before. 
%This shows that when the query keyword didn't show in the training set, the model didn't learn keyword pattern and the performance was poor.
We evaluated the three detectors mentioned in Section~\ref{subsec:key_det}, denoted in  Table~\ref{table:att_exp} as NN, Cos, and NN+Cos.
Regardless of the model, using the network for keyterm detection always works better than simply computing cosine similarity (NN v.s Cos).
Taking cosine similarity as another input to the keyterm detection network (NN+Cos) does not improve performance on test set 1, but does improve the performance in some cases on test sets 2 and 3.

%In comparing normalization methods, %(parts (B) v.s. (C)). 
%we find that smooth attention normalization also outperforms the baselines (part (D) v.s. rows (A), (B)), except for the result in row (D-2).
%However, the sharp attention normalization yields better performance than the smooth version regardless of the keyword detector (parts (C) v.s. (B)). 
%This is reasonable, because the audio segment usually contains only one query term: sharp attention thus yields better performance, as it focuses the network on a narrow span in the audio segment.
Part (D) shows the integration of the DTW output in row (A) and the attention-based model in part (C), for which the integration weight is 0.4 for DTW and 0.6 for the attention-based model. 
We observe improvements for test sets 1 and 3.
This shows that DTW and the attention-based model are complementary.
%This shows that DTW still have some information not captured by attention-based model.
%Bidirectional model has significant advancement in question answering problem. But now the target is matching speech signal,the bidirectional model may not be able to make significant improvement. 

\subsection{Attention analysis}  \label{subsec:attana_exp}

\begin{figure}[!th]
\centering
\includegraphics[scale=0.40]{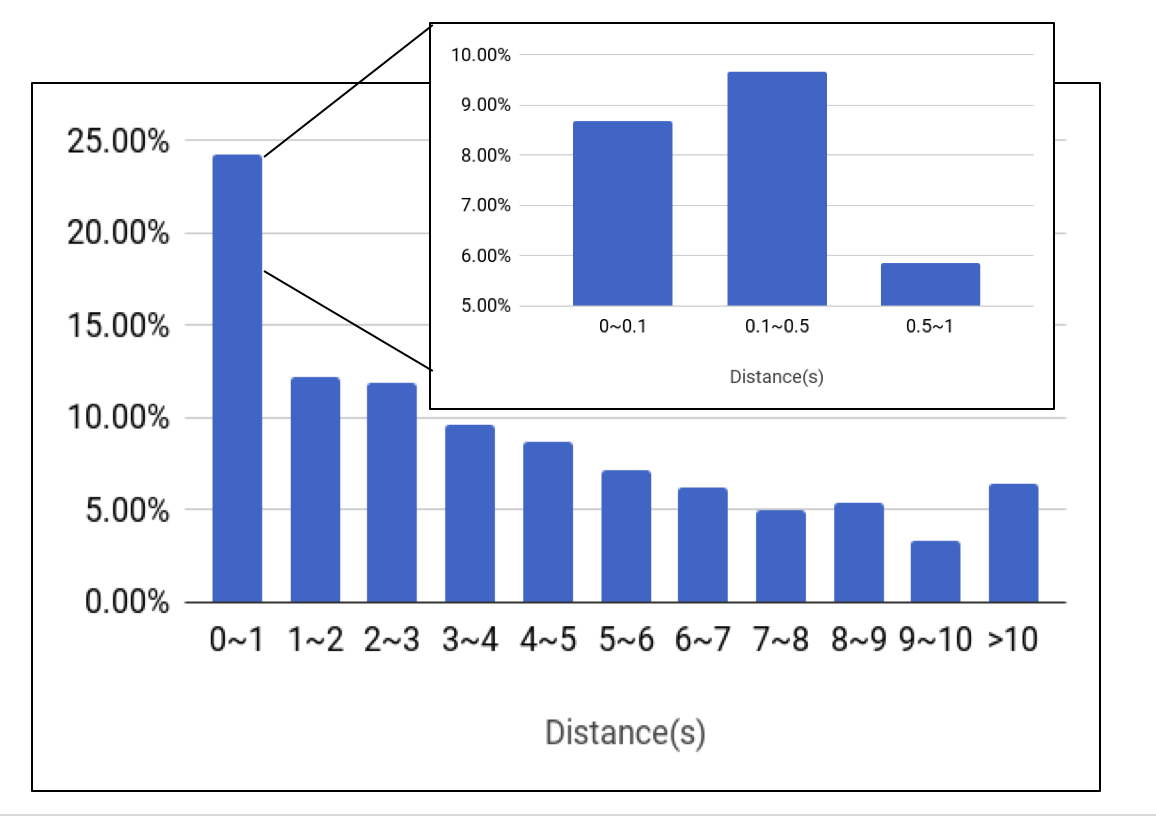} 
\caption{Time differences between maximum attention weight and end of query.}
\label{fig:att_dis}
\vspace{-2mm}
\end{figure}

In spoken term detection, we seek to determine not only whether the query term exists in the audio segments, but sometimes also -- if they exist -- the time spans of these query terms within the segments.
We find that the attention weights reveal the time spans of the query terms.
%Here we try to analysis attention and want to figure out why attention is so helpful. 
In Fig.~\ref{fig:att_dis} we present the analysis of test set 1.
For all query terms and for the audio segments containing the query terms, we compute the time difference between the position with the highest attention weight and the end of the query. 
The horizontal axis in the figure shows the time difference, while the vertical axis is the percentage among all the audio segments considered.
From the figure, we find that distances under one second accounted for 25\% of the cases;
that is, in these cases the attention mechanism located the query term with less than a one-second error. 
Further analysis in the enclosed subfigure shows the time duration from 0 to 1 seconds.
We find that most of the time differences fall between 0.1 and 0.5; this shows that the highest peak of the attention weights are quite close to the query word. %The distance within 0.1 second also account a dominant part,so the highest peak is pretty close to query word.
This suggests that attention yields a precise focus on the end location of the query. 

\subsection{Multiple Hopping} \label{subsec:hop_exp}

\begin{table}[ht]
	 \centering
%	\vspace{-5mm}
    \caption{Multiple-hop results}
	 \label{table:hopping}
    % \resizebox{\columnwidth}{!}{%
	 \begin{tabular}{|c|c|c|c|}
		 \hline
		 & Test set 1 & Test set 2 & Test set 3 \\
		 \hline
		 (A): DTW                    & 0.6173 & 0.5778 & 0.5678 \\
         \hline\hline
         (B): 1-hop                  & 0.6523 & 0.6246 & 0.5754 \\
         \hline
		 (C): 2-hop                  & 0.6472 & 0.6430 & 0.5842 \\
		 \hline
         (D): 3-hop                  & 0.6676 & 0.6404 & 0.5837 \\
         \hline
         (E): 4-hop                  & 0.6417 & 0.6476 & 0.5792 \\
         \hline 
%         (F): 5-hop                  & 0.6593 & 0.6344 & 0.5853 \\
 %        \hline
         \hline
         (F): (A)+(D)                & 0.6789 & 0.6430 & 0.5830 \\
         \hline
	   \end{tabular}%
     %  }
\end{table}

Table~\ref{table:hopping} shows the results when using multiple hops to generate audio segment representations.
The results in row (B) are the results without multiple hops, also shown in row (C-1) of Table~\ref{table:att_exp}; the results with 2 to 4 hops are those in rows (C) to (E).
Multiple hops outperform single hops (rows (C) to (E) v.s. (B)), except for 1 and 3 hops on test set 1.
This shows that hopping improves model generality because in test sets 2 and 3 the training and testing data are mismatched. 
In row (F), we also integrated the DTW and 3-hop results, yielding further improvements to the MAP score on test set 1.

\subsection{Unsupervised Scenario}
%\vspace{-4mm}
\vspace{-5mm}
\label{subsec:un_exp}
\begin{table}[ht]
	 \centering
	 \caption{Results of attention-based multi-hop network learning from a teacher approach (DTW).}
	 \label{table:unsupervise}
  %   \resizebox{\columnwidth}{!}{%
	 \begin{tabular}{|c|c|c|c|}
		 \hline
		 & Test set 1 & Test set 2 & Test set 3 \\
		 \hline
		 (A) DTW & 0.6173 & 0.5778 & 0.5678\\
         \hline
         (B) Attention $+$ 1-hop &0.6128 & 0.5893&0.5548\\
         \hline
         (C) Attention $+$ 3-hop &0.6141 &0.5964 & 0.5702\\
         \hline
	   \end{tabular}%
 %      }
\end{table}

Here the network is learned from a teacher approach, so no extra label data is needed. %without supervision.
We use DTW as the teacher approach, and normalize the DTW similarity scores between 0 and 1 as the target of regression. 
The results are shown in Table~\ref{table:unsupervise}.
From the table, we find that the performance of the attention-based network without multiple hops is comparable to DTW (rows (B) v.s. (A)), and that the 3-hop network outperforms DTW on test sets 2 and 3 (rows (C) v.s. (A)). 
Here we emphasize that the time complexity of the network during testing is far less than DTW:
given a document length of $M$ and a query length of $N$, the time complexity of DTW is $O( M\times N )$, while the time complexity of the network is $O( M \times n )$, where $n$ is the number of hops\footnote{We implemented DTW in C++ and the network using Tensorflow. On average, without using a GPU, the proposed approach was 7 times faster than DTW.}. 
Therefore, it is reasonable to replace DTW with a network learned from it.

%\vspace{-2mm}
\section{Conclusion}
%\vspace{-2mm}
In this paper, we propose an end-to-end query-by-example STD model based on an attention-based multi-hop network. %which has been successfully used in QA.
The model can be trained in either an supervised or unsupervised fashion.
In the supervised scenario, we show that attention and multiple hops are both very helpful, and that the attention weights of the proposed model reveal the time span of the input keyterm.
In the unsupervised setting, the neural network mimics DTW behavior, and achieves performance comparable to DTW with shorter runtimes. 
In the future, we will explore more new attention-based models, and investigate new models which directly output time spans instead of a confidence score. 
We will also compare the performance of the proposed approach and DTW on posterior features and cross-lingual bottleneck features. %are widely used 
%We will also apply the attention-based multi-hop model on STD with text query.

\bibliographystyle{IEEEbib}
\bibliography{strings,refs}

\end{document}